\documentclass[compsoc,conference,a4paper,10pt,times]{IEEEtran}
\IEEEoverridecommandlockouts
\usepackage{cite}
\usepackage{amsmath,amssymb,amsfonts}
\usepackage{graphicx}
\usepackage{textcomp}
\usepackage{bmpsize}
\usepackage{xcolor}
\usepackage{lipsum}
\usepackage{algorithm}
\usepackage{algpseudocode}
\usepackage{amsmath}
\usepackage{booktabs}

\usepackage[colorlinks=true,urlcolor=black]{hyperref}
\def\BibTeX{{\rm B\kern-.05em{\sc i\kern-.025em b}\kern-.08em
    T\kern-.1667em\lower.7ex\hbox{E}\kern-.125emX}}
\begin{document}

\title{Prompted Contextual Vectors for Spear-Phishing Detection}
\author{\IEEEauthorblockN{Daniel Nahmias}
\IEEEauthorblockA{\textit{Ben-Gurion University of the Negev} \\
\textit{Accenture Cyber Research Lab}}
\and
\IEEEauthorblockN{Gal Engelberg}
\IEEEauthorblockA{
\textit{Accenture Cyber Research Lab}}
\and
\IEEEauthorblockN{Dan Klein}
\IEEEauthorblockA{
\textit{Accenture Cyber Research Lab}}
\and
\IEEEauthorblockN{Asaf Shabtai}
\IEEEauthorblockA{
\textit{Ben-Gurion University of the Negev}}
}

\maketitle

\begin{abstract}
Spear-phishing attacks present a significant security challenge, with large language models (LLMs) escalating the threat by generating convincing emails and facilitating target reconnaissance. 
To address this, we propose a detection approach based on a novel document vectorization method that utilizes an ensemble of LLMs to create representation vectors. 
By prompting LLMs to reason and respond to human-crafted questions, we quantify the presence of common persuasion principles in the email's content, producing prompted contextual document vectors for a downstream supervised machine learning model.  
We evaluate our method using a unique dataset generated by a proprietary system that automates target reconnaissance and spear-phishing email creation. 
Our method achieves a 91\% F1 score in identifying LLM-generated spear-phishing emails, with the training set comprising only traditional phishing and benign emails. 
Key contributions include a novel document vectorization method utilizing LLM reasoning, a publicly available dataset of high-quality spear-phishing emails, and the demonstrated effectiveness of our method in detecting such emails. 
This methodology can be utilized for various document classification tasks, particularly in adversarial problem domains.
\end{abstract}

\begin{IEEEkeywords}
Spear-phishing, GenAI, Machine Learning, Concept Drift, Social Engineering
\end{IEEEkeywords}

\section{\label{sec:intro}Introduction}

Spear-phishing, a sophisticated variant of phishing attacks~\cite{checkpoint_spearphishing}, targets specific individuals or organizations using malicious emails or other types of messages. 
The primary objectives of spear-phishing are to acquire sensitive information, such as login credentials, or to compromise the targets' device by installing malicious software (e.g., by clicking on a link)~\cite{forescout_spear_phsihing} on the device.

Distinguished by its personalized approach, spear-phishing relies on extensive research and data collection on the targeted individuals or organizations. 
This thorough investigation enables attackers to craft deceptive emails that appear to originate from trusted sources within the targets' personal or professional networks~\cite{spear_phishing_in_a_barrel}. 
In essence, spear-phishing prioritizes the quality of engagements over the sheer volume of attempts seen in more generic phishing methods.
The considerable investment of time and effort researching and tailoring messages so they appear as if they originate from legitimate sources enhances the likelihood of deceiving potential victims. 
This was highlighted in a recent report by Barracuda~\cite{barracuda_report} that reports that although spear-phishing attacks represent just 0.1\% of all email-based attacks, they account for 66\% of the security breaches resulting from email-based attacks.

Large language models (LLMs) are transforming the spear-phishing threat landscape~\cite{hazell2023large}, influencing it in two significant ways: 
\begin{enumerate}
    \item LLMs facilitate the automated creation of sophisticated spear-phishing emails, enhancing the overall quality of deceptive communication~\cite{heiding2023devising,elatoubi2023phishing}. This capability streamlines the attackers' ability to craft and rapidly disseminate convincing messages tailored to specific targets.
    \item The integration of LLM-powered agents into the target reconnaissance process enhances efficiency~\cite{llm_recon,zvelo_llm_recon}, automating a previously time-consuming and manual task performed by humans.
\end{enumerate}

This advancement in intelligence gathering not only streamlines the attackers' operations but also amplifies the scale at which potential vulnerabilities can be identified and exploited~\cite{llm_recon,yao2023survey}.
Moreover, the Federal Bureau of Investigation (FBI) has confirmed that attackers are already using AI-based tools, both publicly available and custom-made, to carry out highly targeted phishing campaigns~\cite{fbi_ai_phishing_warning}. 
This calls for suitable detection techniques, particularly considering that spear-phishing detection is already a challenging problem to address~\cite{kaspersky_spear_phishing}. 

Previous studies aimed at detecting spear-phishing attacks~\cite{gascon2018reading,khonji2011mitigation,Duman2016,Stringhini2014} have diverged from the conventional classification problem of distinguishing between legitimate and malicious emails, as often seen in studies aimed at detecting traditional phishing attacks. 
A prevalent strategy in spear-phishing detection studies has been to adopt an attribution approach, which seeks to classify an email within the context of a known spear-phishing campaign or a trusted author~\cite{das2019sok}. 
While this approach can prove effective, its utility is restricted by the requirement for ground-truth samples from known spear-phishing campaigns (or trusted authors), thereby limiting its scalability and effectiveness in detecting unseen campaigns. 

We posit that this shift to the attribution approach stems from the inherent difficulty in distinguishing between spear-phishing emails and legitimate ones.
Moreover, the studies mentioned above rely on traditional feature extraction techniques (e.g., specific word counts, attachment size, body length, number of recipients) 
These kind of features are susceptible to concept drift~\cite{gutierrez2018learning,salloum2021phishing}, and their explainability is limited.
Another limitation of these studies is that the spear-phishing emails used for evaluation have not been published. 
This makes it difficult for other researchers to evaluate their proposed detection methods on spear-phishing emails employed in previous studies.
To facilitate future research in this field, we publish the spear-phishing dataset we evaluated our detection method on. 

In this paper, we introduce a novel document vectorization method, which we use for the challenging task of detecting spear-phishing attacks.
Our proposed vectorization method utilizes an ensemble of LLMs, i.e., a LLM-based committee of experts, and prompts each of them to reason about human-crafted questions regarding the contextual information of the document's content (in our case, the inspected email.)
The questions were crafted to detect common persuasion techniques often employed in social engineering attacks. 
These techniques are derived from studied persuasion principals~\cite{cialdini2001science,ferreira2019persuasion}, exploiting human vulnerabilities to persuade the target in taking a specific action which benefits the attacker.
An example of such a persuasion principle is the authority principle, which attackers often employ a technique derived from it - impersonating an authoritative figure and asking the target to perform their urgent request as soon as possible. 

The LLMs are then prompted to quantify their answers to these questions as a probability for each email in our dataset.
By concatenating the probabilities generated by each LLM in the ensemble for every question regarding the email, an explainable representation vector is created for every email.
These vectors, which we refer to as \textit{"prompted contextual vectors,"} serve as input for a supervised machine learning (ML) model used for the task of spear-phishing detection. 
Our detection approach harnesses the unique capability of LLMs to understand context and reason about questions regarding an email's content, a functionality not possessed by existing phishing detection methods, which mainly rely on traditional natural language processing (NLP) techniques~\cite{phishing_nlp}.

To evaluate our proposed method, we employ a unique dataset produced using a proprietary system developed by a collaborating company.\footnote{The name of the company is not disclosed in order to maintain the anonymity of the authors} 
This system automates both the target reconnaissance process and the generation of spear-phishing emails. 
This automated system allowed us to produce a set of 333 high-quality spear-phishing emails grounded in information collected by the system's reconnaissance agent. 
The dataset is designed to emulate the characteristics of the next generation of spear-phishing campaigns.
We augment the dataset with legitimate emails from the Enron~\cite{klimt2004enron} email dataset and both legitimate and traditional phishing emails from the SpamAssasin~\cite{spamassasin} dataset. 
Both datasets are publicly available. 
We compared our method against multiple state-of-the-art document vectorization methods based on the transformer architecture~\cite{attention}.
The results demonstrate that our proposed document vectorization method, which powers the downstream phishing detector, generates more informative representational vectors. This allows even a straightforward ML model like $k$-nearest neighbors to effectively identify LLM-generated spear-phishing emails, achieving an F1 score of 91\%.
The same classifier demonstrated a significantly lower F1 score when using the existing document vectorization methods examined.
In our experiments, the malicious class within the training dataset consisted exclusively of traditional phishing emails, with spear-phishing emails appearing only in the test dataset. This setup naturally simulates concept drift~\cite{concept_drift2016}, capturing the evolution of malicious content from simple generic scams to sophisticated, AI-powered spear-phishing campaigns.
To the best of our knowledge, no other phishing or spear-phishing detection study has evaluated their proposed methods under such conditions.
Furthermore, we assess our method's capabilities of generalizing to other forms of social engineering attacks by conducting an experiment in which the training dataset comprises emails, while the test dataset exclusively contains smishing (SMS phishing) and benign SMS messages. 
In this experiment, our method achieved a 90\% F1 score.

The main contributions of this paper are:
\begin{enumerate}
    \item We present a novel document vectorization method, leveraging LLMs' reasoning capabilities to produce prompted contextual document vectors.
    While we employ this technique for the task of spear-phishing detection, it may be useful for other document classification tasks, especially in adversarial problem domains.
    \item We introduce and publish a unique dataset consisting of high-quality spear-phishing emails generated by an LLM powered system. To the best of our knowledge, there is currently no publicly available dataset of this type. 
    \item We demonstrate the effectiveness of prompted contextual vectors as input to a detection pipeline, successfully identifying spear-phishing emails even when they're not in the classifier's training set.
    \item We demonstrate that using prompted contextual vectors as input to the detection pipeline enables it to generalize to other forms of social engineering attacks, successfully identifying novel threats like smishing messages, even though they were absent from the training data. 
\end{enumerate}
\section{\label{sec:relatedworks}Related Work}

This paper proposes a novel document vectorization method that leverages LLMs' reasoning capabilities for the challenging task of spear-phishing detection. In this section, we survey studies in the field of phishing detection, spear-phishing detection, and feature extraction using LLMs.

\subsection{Phishing Detection}\label{subsec:phishing_related}
Recent studies on phishing detection have proposed various methods aimed at improving the detection accuracy and reducing false positives. 
Bountakas et al.~\cite{bountakas2023helphed} introduced HELPHED, a novel hybrid ensemble learning approach for phishing email detection. 
The proposed method combines stacking and soft voting ensemble learning techniques with ML algorithms to process emails' content (e.g., count of attachments, images and URLs) and text (e.g., count of specific words) features. This work builds on previous work in this field by incorporating features suggested in earlier phishing detection studies~\cite{akinyelu2014classification,smadi2015detection,ma2009detecting,anandita2017novel,a2011hybrid,toolan2009phishing} when suggesting new features. 
The HELPHED approach achieved a high F1-score of 0.994.

In a similar vein, Alhogail et al.~\cite{alhogail2021applying} proposed a deep learning-based classifier using a graph convolutional network (GCN) and traditional NLP methods for phishing email detection. Their method includes three phases: data preprocessing (email cleaning, tokenization, and removal of stop and rare words), graph construction (building a single large graph from the email corpus with words and emails as nodes and co-occurrences as edges), and classification (training the GCN to distinguish between phishing and ham emails). The classifier achieved 98.2\% accuracy, 98.5\% precision, and 98.3\% recall.

In another study~\cite{d-fence}, the authors proposed a method using deep learning and ensemble learning to analyze three email modules: structure, text, and URL. Each module has its own classifier. The structure module, which uses a tree-based classifier, focuses on features like the number of links and domains. The text module uses BERT embeddings for the raw text and trains a classifier on these embeddings. The URL module classifies URLs with a CNN-LSTM architecture. These module predictions are then combined by a meta-classifier to determine if an email is malicious or benign. The method achieved a 95\% recall score on both private and public datasets.

These studies drive progress in phishing detection by presenting novel methodologies, utilizing ensemble learning, deep learning, and graph-based analysis. However, their reliance on traditional NLP feature extraction techniques, like word counts, may be less effective in detecting personalized spear-phishing emails.
None of the abovementioned studies tested their detection methods on spear-phishing emails.

\subsection{Spear-Phishing Detection}
The field of spear-phishing detection has seen relatively fewer studies compared to phishing detection.
Given the complex nature of spear-phishing detection, the majority of studies utilizing ML for this purpose have adopted the attribution approach~\cite{evans2022raider}. 
This involves attributing an email either to a known spear-phishing campaign or a recognized sender.
The attribution approach has significant limitations, because it relies on the assumption that attackers will always impersonate a known sender. However, this is not always the case~\cite{imperva_spear_phsihing}. Attackers can perform spear-phishing attacks by learning about their target and use the information gathered in order to impersonate an external business contact. In these cases, the attribution approach will fail.

Gascon et al.~\cite{gascon2018reading} presented a method for detecting spear-phishing emails that assesses whether an email is attributed to a known sender based on the structural traits of an email (e.g., number of attached files, number of cc's and bcc's), rather than its content.
The proposed method learns profiles for a large set of senders and identifies spoofed emails as deviations from these profiles.
The method is based on the observation that a sender leaves characteristic traits in the structure of an email, which are independent from textual content and often persist over time. 
These traits significantly differ between senders and reflect peculiarities of the user's behavior, email client, and delivery path. 

Evans et al.~\cite{evans2022raider} built on the work of Gascon et al. ~\cite{gascon2018reading}, utilizing reinforcement learning to identify the most significant features proposed in the prior study. The results suggested that using reinforcement learning to automatically identify significant features could reduce the dimensions of the required features by 55\% while obtaining comparable results. Evans et al. formulated their experiments as a binary classification task and employed a spear-phishing dataset that they manually curated by altering benign emails within their dataset. The authors emphasized the necessity of including a minimum of two emails from a known sender, with their messages modified to generate new spear-phishing samples in the training dataset for their recommended approach to be effective. Therefore, it could be said that this study also takes the attribution approach.

Stringhini el al.~\cite{Stringhini2014} proposed a system designed to validate email authorship by learning users' typical email-sending behaviors, such as word and character counts, over time and comparing subsequent emails sent from their accounts against this model. Their findings revealed that the system is capable of detecting spear-phishing emails with a detection rate exceeding 90\% by matching the emails against known users, provided that the users have a sending history of at least 1,000 emails.
However, this method is likely to fail if the attacker can craft an email that closely mimics the legitimate user's email-sending behavior.

Han et al.~\cite{han2016accurate} also performed  a spear-phishing attribution study to link suspected emails to known campaigns. They used a semi-supervised learning model to attribute emails to campaigns and detect new ones. The model extracted four categories of email features: origin (e.g., domain, IP), text (e.g., word counts, topics), attachment (size, type), and recipient (e.g., domain, organization). Using an affinity graph-based model, they achieved a 0.9 F1 score with a 1\% false positive rate for known campaigns. The authors also conducted a spear-phishing detection (binary classification) experiment, in which their proposed email representational vector served as input for a random forest classifier. The results were notable, with the classifier achieving recall score of 95\%, while using only 6\% of the labeled data as the training set. It is essential to acknowledge, however, that the dataset consisted of six known spear-phishing campaigns, likely sharing common features. Given the relatively basic nature of the suggested features, this method is susceptible to concept drift and may fail to address attacks from unknown campaigns. For instance, an attack originating from a country or IP address not encountered by the model before could lead to failure.

Ding et al.~\cite{ding2021spear} presented a spear-phishing detection method in which 21 stylometric features were extracted from email logs, three forwarding features were extracted from the Email Forwarding Relationship Graph Database (EFRGD), and three reputation features were extracted from two third-party threat intelligence platforms: VirusTotal and Phish Tank. 
The authors evaluated their method on 417 spear-phishing emails collected by security experts. 
The proposed method obtained a recall score of 95.56\% and a precision score of 98.85\%. 
The main limitation of the approach is that it relies on features provided by external vendors which might be inaccessible for a variety of reasons. In contrast, our approach relies solely on the body of the email it needs to classify. 

It is notable that among the abovementioned studies, only two (\cite{ding2021spear,han2016accurate}) proposed a spear-phishing detection method instead of exclusively adopting the attribution approach. 
A primary limitation shared by these studies is their reliance on basic features, which are susceptible to concept drift and offer limited explainability.

\subsection{Feature Extraction Using LLMs}
LLMs and their use in downstream ML tasks is a relatively new field of study. 
Several studies have proposed feature extraction approaches that share similarities to the document vectorization approach we suggest. 
In~\cite{mcinerney2023chill}, the authors used the Flan-T5 LLM to vectorize patient notes for classification tasks by prompting it with yes/no questions and using the presence of ‘yes’ in the response as a binary feature, or a continuous representation based on the ratio of ‘yes’ to ‘no.’ Their evaluation showed that these features performed worse than TF-IDF vectors but were valued for their interpretability in clinical settings.
Our study differs by using chain-of-thoughts (CoT) prompting~\cite{wei2022chain} for continuous feature generation directly from the LLM. We also employ an ensemble of LLMs for robust vectorization and evaluate our method on an adversarial task, unlike the task of patient notes classification.

In~\cite{wu2023cheap}, Wu et al. used GPT-3.5 to generate feature vectors for 'cheapfakes' detection in news articles, instructing the model to produce a single list of integers (0-9) based on predefined questions about sentence relationships. This improved their existing classifier's accuracy by 0.55 on their test dataset. While both studies use LLMs for text vectorization in adversarial settings, our approach differs significantly. We prompt an ensemble of LLMs to answer each predefined question separately, incorporating Chain-of-Thought reasoning for each response. This contrasts with Wu et al.'s method of asking a single LLM to output an integer list covering all questions in one prompt without explicit reasoning. By quantifying each reasoned response as a probability for the corresponding vector element, our method aims to produce more nuanced and reliable results.
\section{\label{sec:background}Background}
\subsection{\label{subsec:concept_drift}Concept Drift in Cyber Security}
In the field of cyber security, machine learning models are increasingly used to detect and prevent various types of cyber threats, such as phishing and malware. However, one of the key challenges these models face is the phenomenon of \textit{dataset shift}~\cite{allix2015your,barbero2022transcending}, which occurs when the statistical properties of the data used for training begin to change over time. In social engineering attacks, this drift is particularly pronounced, as adversaries continuously adapt their tactics to bypass detection systems~\cite{cyberscoop_nlp_evasion}. As a result, the assumption that the distribution of data remains identical between training and test phases is violated, leading to a decline in the model's ability to generalize to new, unseen threats. 

Dataset shift can be categorized into the following types~\cite{moreno2012unifying}:

    \begin{itemize}
        \item \textbf{Covariate Shift~\cite{covariate_shift}}: This occurs when the distribution of features in the input data \(P(x \in X)\) (e.g., email structure, content, or sender information) changes over time. For instance, phishing emails may shift from using simple scams to leveraging AI-generated content that mimics legitimate communication more closely. In malware detection, this could happen when new variants of malware begin using different evasion techniques. For example, early malware might have relied heavily on certain API calls, but over time, more sophisticated malware might start using different API calls to avoid detection.
        \item \textbf{Label Shift~\cite{label_shift}}: This involves changes in the distribution of labels \(P(y \in Y)\) (e.g., the prevalence of certain attack types). For example, an increase in targeted spear-phishing attacks may alter the overall distribution of phishing types within a dataset, affecting the model’s ability to classify emails correctly.
        \item \textbf{Concept Drift~\cite{widmer_concept_drift}}: This refers to changes in the conditional distribution \(P(y \in Y | x \in X)\), where the relationship between the input features and their corresponding labels evolves. In social engineering detection, this could occur when new types of social engineering attacks, such as smishing (SMS phishing), emerge, and are not adequately captured by features learned from traditional phishing data.
    \end{itemize}

In cyber-security research, it is common to refer to all forms of distributional changes collectively as concept drift, given the difficulty in precisely attributing model performance degradation to specific types of shifts~\cite{moreno2012unifying,barbero2022transcending}. 
In this study, we will continue to use the term 'concept drift' as a general term encompassing all three categories of dataset shift.

Addressing concept drift is essential to ensuring that detection systems can continue to identify both existing and emerging threats in an ever-evolving threat landscape.

There are several approaches to addressing concept drift. One common method is to detect drift and reject predictions where drift is suspected~\cite{barbero2022transcending}. However, a more effective strategy is to design a feature space that aligns with the theorized optimal feature space $X$, which captures the core malicious behaviors of a threat class and remains robust across different types of dataset shifts. This is the approach this work aims to implement.
\section{\label{sec:method}Methodology}

\begin{figure*}[t]
    \centering
    \includegraphics[width=0.8\linewidth]{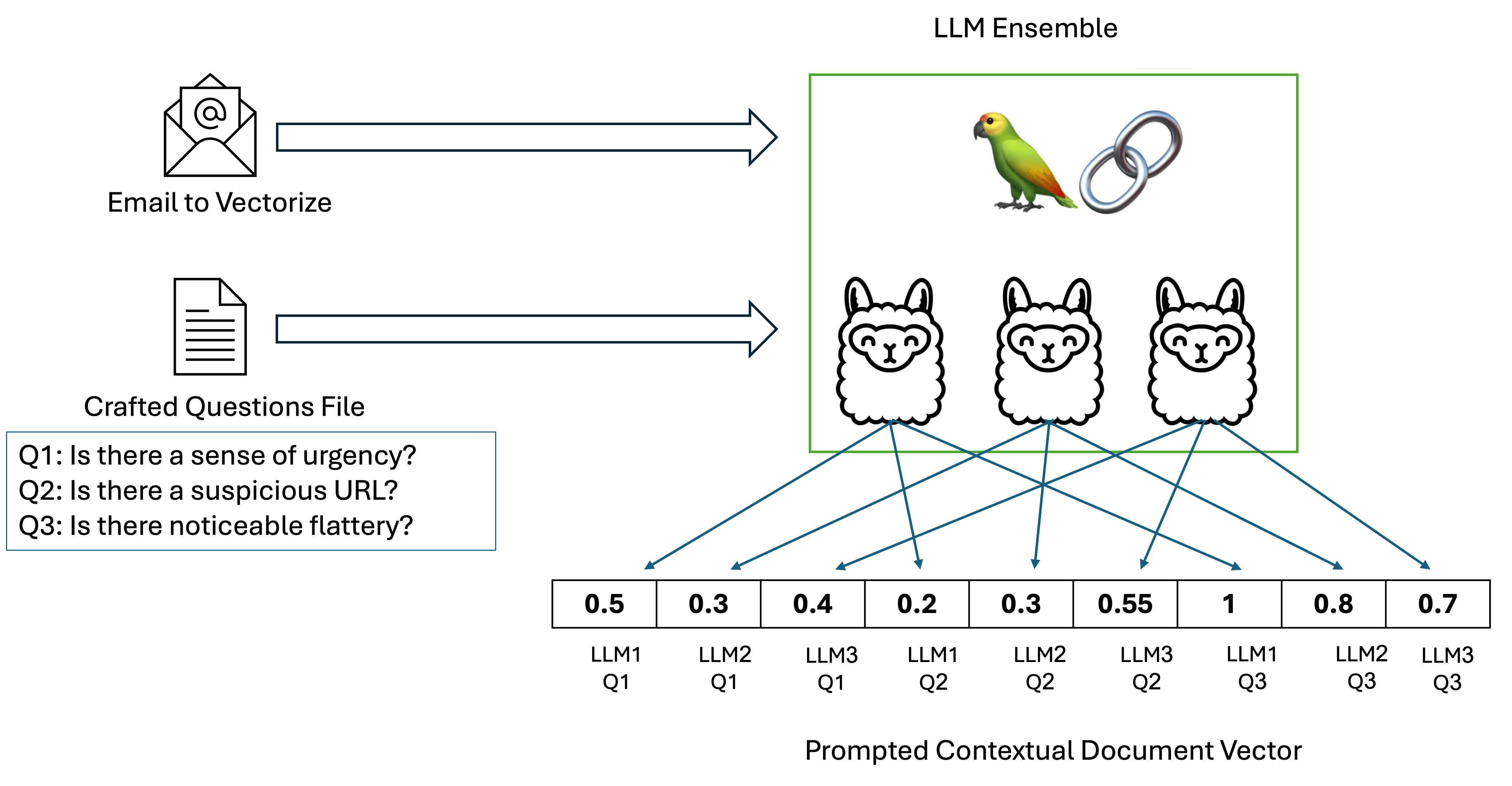}
    \caption{Prompted contextual vectorization process.}
    \label{fig:vectorization_method}
\end{figure*}

\subsection{Method Overview} 
In this paper, we introduce a novel method for document vectorization that leverages the contextual reasoning capabilities of LLMs. 
By utilizing an ensemble of LLMs, each element of the generated vector is derived through responses to a series of human-crafted questions, tailored for a specific task. 
Our vectorization method is applied and examined in the scope of spear-phishing detection.

The overall pipeline, illustrated in Figure~\ref{fig:vectorization_method}, utilizes our method to vectorize a specific email. 
The pipeline involves extracting the textual content of the email, feeding it as input, along with the set of questions, to the ensemble of LLMs, and obtaining a numerical vector that represents the email. 
This vector can then be employed by downstream machine learning tasks. 

The algorithmic flow of the dataset vectorization pipeline is illustrated in the pseudocode outlined in Algorithm~\ref{dataset_creation_alg}:
Initially, we create a new array to depict the dataset (line 1). Then, we proceed to iterate through each email, generating a new array to represent the email's vector (lines 2-3). Within this loop, we iterate through every possible combination of an LLM (in the ensemble) and a question from the questions set. The question is presented to the LLM along with the inspected email and its response is appended into the array representing the email vector (lines 4-9). Eventually, after iterating through all possible combinations of questions and LLMs, this vector is appended to the array signifying the dataset and we proceed to the next email. Upon completing this procedure for all emails, we obtain a vector dataset that can be used in any downstream ML task.
In the subsections below, we describe the components of our proposed vectorization method, exploring the inherent properties of the vectors generated by our method, highlighting the benefits of employing an ensemble of LLMs, and examining the process of crafting the questions relevant to our task of detecting spear-phishing emails.

\subsection{Prompted Contextual Document Vectors}

Chat or instruction-aligned LLMs possess a novel capability lacking in traditional models: the ability to generate features informed by the contextual nuances of the text provided. 
We leverage this capability to represent a given email (i.e., document) as a numerical vector. 
In essence, we prompt the LLM to evaluate the extent to which a particular aspect of the email content is present in a given document (i.e., quantify the presence of the aspect in the document), requesting the LLM to produce a floating-point number that represents this quantitatively. 
By prompting an LLM with a series of distinct questions, we can obtain a vector of floating-point numbers, where each element in the vector quantifies the presence of a specific aspect in the text, based on a human-crafted question. 

We propose a new term for vectors generated by our document vectorization method: \textit{prompted contextual document vectors}.

Current content-based phishing detection solutions rely on basic NLP features such as specific word counts, detecting spelling errors, or context-less word embeddings.
We hypothesize that these features will prove unreliable for detecting sophisticated spear-phishing emails generated by LLMs. Therefore, we designed the prompted contextual vectors to capture malicious intent embedded in the email's content, rather than stylometric features. Our approach aims to align the feature space more closely with a theorized optimal feature space $X$, thereby achieving greater robustness against various forms of dataset shift that may arise from evolving attack patterns.

\subsection{Ensemble of LLMs for Robust Prompted Contextual Document Vectors}

We enhance the robustness of our proposed vectorization method by employing an ensemble of LLMs instead of relying on a single LLM which may be susceptible to hallucinations or biases inherent in the data it was trained or fine-tuned on. 
Similarly to widely-used ensemble ML techniques~\cite{sagi2018ensemble}, our goal is to introduce randomness into the learning process in order to avoid overfitting. 
To accomplish this, we extract prompted contextual features from multiple LLMs, each built using a unique transformer architecture and likely trained on different datasets. 
By including different 'opinions' derived from a diverse set of LLMs as features, we inject variability into the document vector. 
Thus, in a departure from conventional ensemble ML methods, our method introduces randomness to the input vector instead of incorporating it in the training process of the primary learner.
The ensemble of LLMs used in our experiments consists of GPT-3.5, GPT-4, and Gemini Pro~\cite{gemini}. 
It is important to note that we use chat-aligned LLMs, as they are easier to prompt for this kind of task.
The temperature parameter of each LLM in the ensemble is set to zero in order to ensure output consistency.

\subsection{\label{subsec:questions}Human Crafted Questions for Prompted Contextual Vectors}
Our goal is to leverage the LLMs' ability to reason regarding the email's content and use it to generate vectors that will then be provided as input to an ML classifier.
To do so, we craft questions that are meaningful for the task of spear-phishing detection, targeting specific aspects of the email's content. These questions are incorporated into prompts that instruct the LLM to reason about them and provide a final answer in a structured output.

We utilize the chain-of-thought (CoT) prompting framework~\cite{wei2022chain}, instructing the models to think step-by-step. The structured output includes both the LLM's reasoning process and a resulting probability, quantifying the likelihood and extent to which the aspect is present in the email.

\begin{algorithm}[t]
\caption{Constructing Vector Dataset from Emails}\label{dataset_creation_alg}
\textbf{Input:} A dataset of emails $E$, a list of questions $Q$, and a list of LLMs $L$ \\
\textbf{Output:} A vectorized dataset of emails

\begin{algorithmic}[1]
\State $ \text{dataset} = []$

\For {every email $e \in E$}
    \State $ \text{email\_vector} = []$

    \For {every question $q \in Q$}

        \For {every LLM $l \in L$}

            \State $ \text{probability} = \text{ask\_question}(e, q, l)$ 

            \State $ \text{email\_vector.append(probability)}$

        \EndFor

    \EndFor

    \State $ \text{dataset.append(email\_vector)}$

\EndFor

\end{algorithmic}
\end{algorithm}

The questions were crafted after examining research papers that investigated the persuasion tactics commonly employed in phishing campaigns.
Specifically, Cialdini et al.~\cite{cialdini2001science} formalized six basic tactics: \textit{reciprocity}, \textit{consistency}, \textit{social proof}, \textit{likeability}, \textit{authority}, and \textit{scarcity}. 
Numerous studies explored the frequency of these tactics in phishing campaigns~\cite{bustio2024uncovering,akbar2014analysing,butavicius2016breaching,ferreira2019persuasion,zielinska2016persuasive}. 
These studies indicate that the principles of social proof, authority, likeability, and scarcity are utilized more frequently compared to others. Therefore, we chose to focus on these specific principles.
\begin{itemize}
    \item \textbf{Social proof} indicates that individuals are likelier to follow through with a request when they see that others have already complied.
    \item \textbf{Scarcity} revolves around the notion that people tend to attribute greater value to items or opportunities that are scarce or limited. This principle holds true even when the limited resource is time, causing individuals to make decisions with less deliberation when faced with time constraints.
    \item \textbf{Authority} suggests that individuals are more inclined to adhere to a request if it emanates from a figure of authority.
    \item \textbf{Likeability} suggest that building trust will in turn increase the likelihood of individuals agreeing to a request. This principle emphasizes the natural tendency for people to trust those they consider attractive and credible.

\end{itemize}

However, those studies primarily focused on analyzing traditional phishing emails, since they are currently more prevalent than spear-phishing emails. 
Therefore, while we create questions to gauge the presence of the abovementioned persuasion tactics, we also generate additional questions to measure new aspects of email content identified in the spear-phishing emails produced by the proprietary system. 
These aspects are likely to be found in spear-phishing emails, whether generated by a language model or not. 
For instance, we specifically instruct the model to determine if the sender addresses the recipient by name and includes suspiciously specific details. 
We also include questions related to the email content, such questions that assess whether an email resembles a marketing email and check for the presence of suspicious links.
Below is the list of questions we crafted for the task of spear-phishing detection:
\begin{itemize}
    \item \textit{Does this email convey a sense of urgency?} (Scarcity)
    \item \textit{Is there a significant amount of flattery evident in the email?} (Likeability)
    \item \textit{Is there a link in this email that appears to be suspicious?}
    \item \textit{Does this email look like a marketing email?}
    \item \textit{Does the email address the recipient by name and with suspiciously specific details?}
    \item \textit{Are there threats of consequences if the recipient does not act immediately?} (Authority)
    \item \textit{Does the email ask the recipient to update an account information or sign a document through a link?}
\end{itemize}

\noindent We acknowledge that a positive response to a single question does not automatically signify a phishing or spear-phishing email. 
Nevertheless, we propose that specific combinations of LLM responses to a subset of these questions may indicate a phishing pattern.

\subsection{Implementation Details}
The proposed vectorization method was developed in Python using the LangChain~\cite{langchain} framework.
All LLMs in the ensemble were instantiated with their default parameters except for the temperature parameter which is set to zero to ensure output consistency.
The link to the code repository is provided in appendix \ref{sec:availability}.

\subsection{Email Classification}
As mentioned earlier, the prompted contextual document vectors can serve as input for any downstream ML task, regardless of the specific task. 
In this paper, which is focused on spear-phishing detection, we evaluate the usefulness of these prompted contextual vectors for a binary classification task (phishing or benign). 
This distinguish our detection approach from prior research in this area, which mainly opts for the attribution approach. 
We employ a simple ML classifier for this purpose, focusing on demonstrating the efficacy of the vectors instead of the classifier's learning process.
\section{Evaluation}

In this section, we evaluate the effectiveness of the prompted contextual vectors produced by our method for the task of spear-phishing detection. 
We compare the prompted contextual vectors to those produced by other state-of-the-art document vectorization methods.
Unlike previous research in the spear-phishing detection domain, we assess our approach as a binary classification problem (ham/phishing) rather than a multi-class author/campaign attribution problem. 
Our goal is to examine the efficacy of our proposed vectorization method in detecting novel spear-phishing campaigns, based on the assumption that such campaigns, despite their sophistication, will inherently employ common persuasion techniques when creating their content.

\subsection{Data}

\subsubsection{Automatic Spear Phishing Generator}
\begin{figure}[h]
    \centering
    \includegraphics[width=0.48\textwidth]{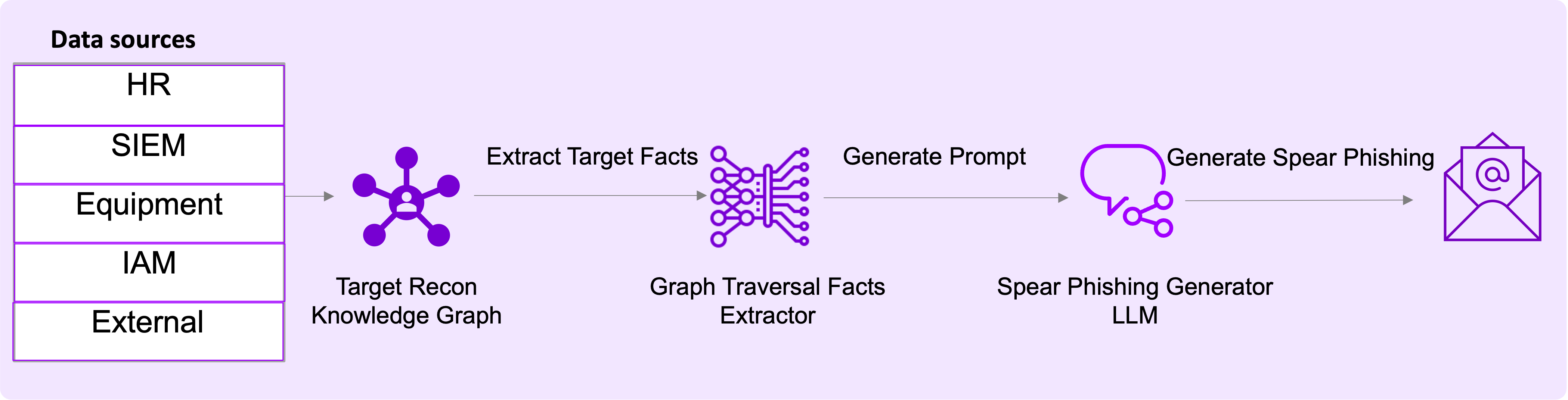}
    \caption{Automatic spear-phishing generator architecture}
    \label{fig:aspg}
\end{figure}

\begin{figure}[h]
    \centering
    \includegraphics[width=0.48\textwidth]{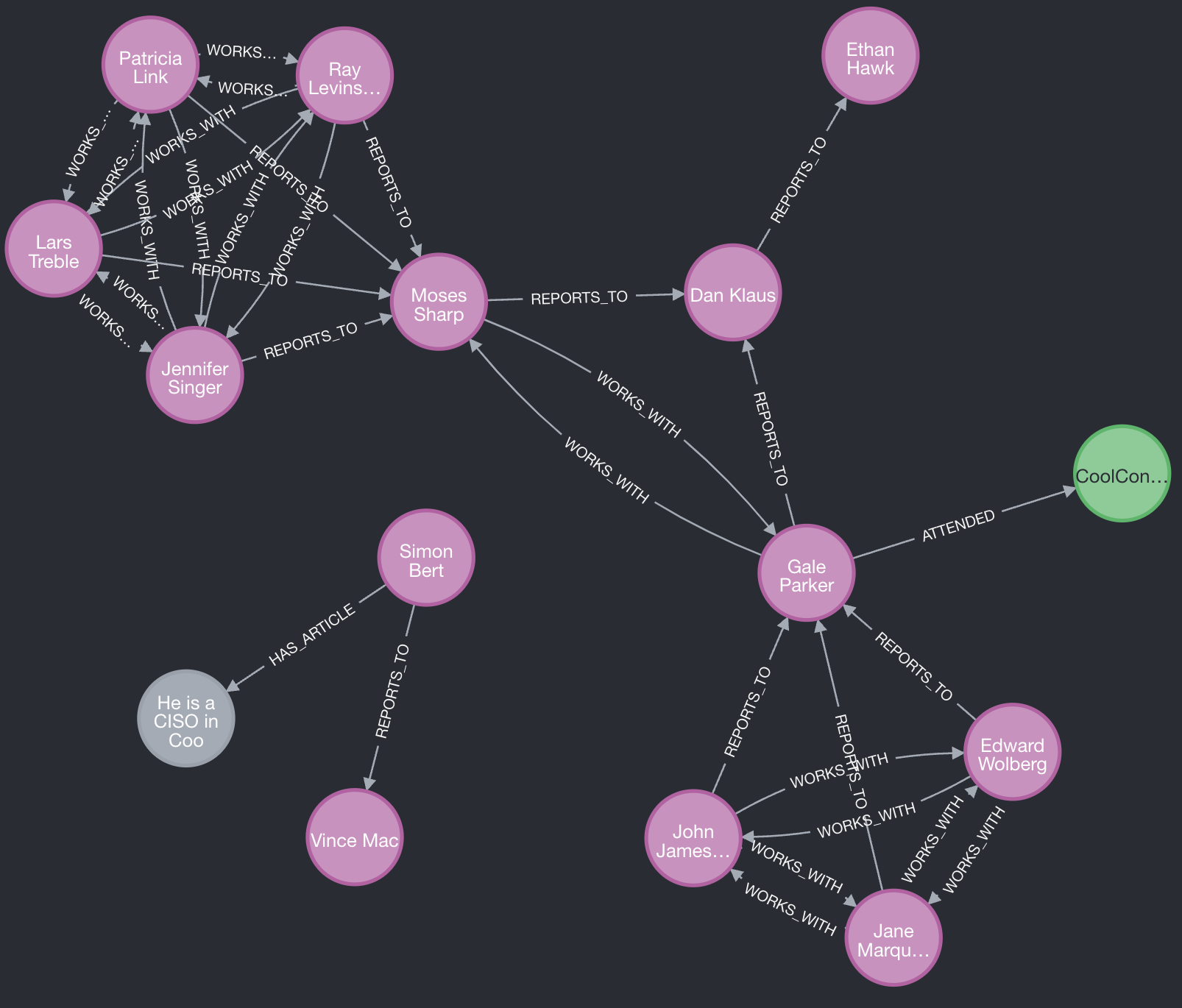}
    \caption{Automatic spear-phishing generator knowledge graph example (anonymized)}
    \label{fig:aspg_graph}
\end{figure}

This subsection describes the automated spear-phishing generation system used to create our dataset. 
Designed to address shortcomings in existing phishing awareness training, the system aims to equip employees with the skills to identify and avoid modern, large-scale LLM-generated spear-phishing attempts.
The system leverages a knowledge graph built from HR, IAM and Equipment Management data, creating a digital twin of the organization with employee hierarchy, access permissions, and work device details.
The system can be augmented with additional data sources, improving the quality and credibility of the generated content. An example of such supplementary information is open-source intelligence (OSINT) data gathered by an autonomous agent. Figure~\ref{fig:aspg_graph} shows a sample of this knowledge graph.
Graph traversal algorithms navigate the knowledge graph to retrieve information regarding every employee. 
This information is then fed to an LLM, which generates a pretext used to craft personalized spear-phishing messages. Figure~\ref{fig:aspg} shows an overview of the system's architecture.
The comprehensive organizational data contained within the system's knowledge graph effectively simulates the intelligence that well-equipped attackers could gather using advanced reconnaissance tools. This approach is particularly relevant given that threat actors are already leveraging both public and proprietary GenAI tools for target reconnaissance and email crafting~\cite{fbi_ai_phishing_warning}.

\begin{figure}[h]
    \centering
    \includegraphics[width=1\linewidth]{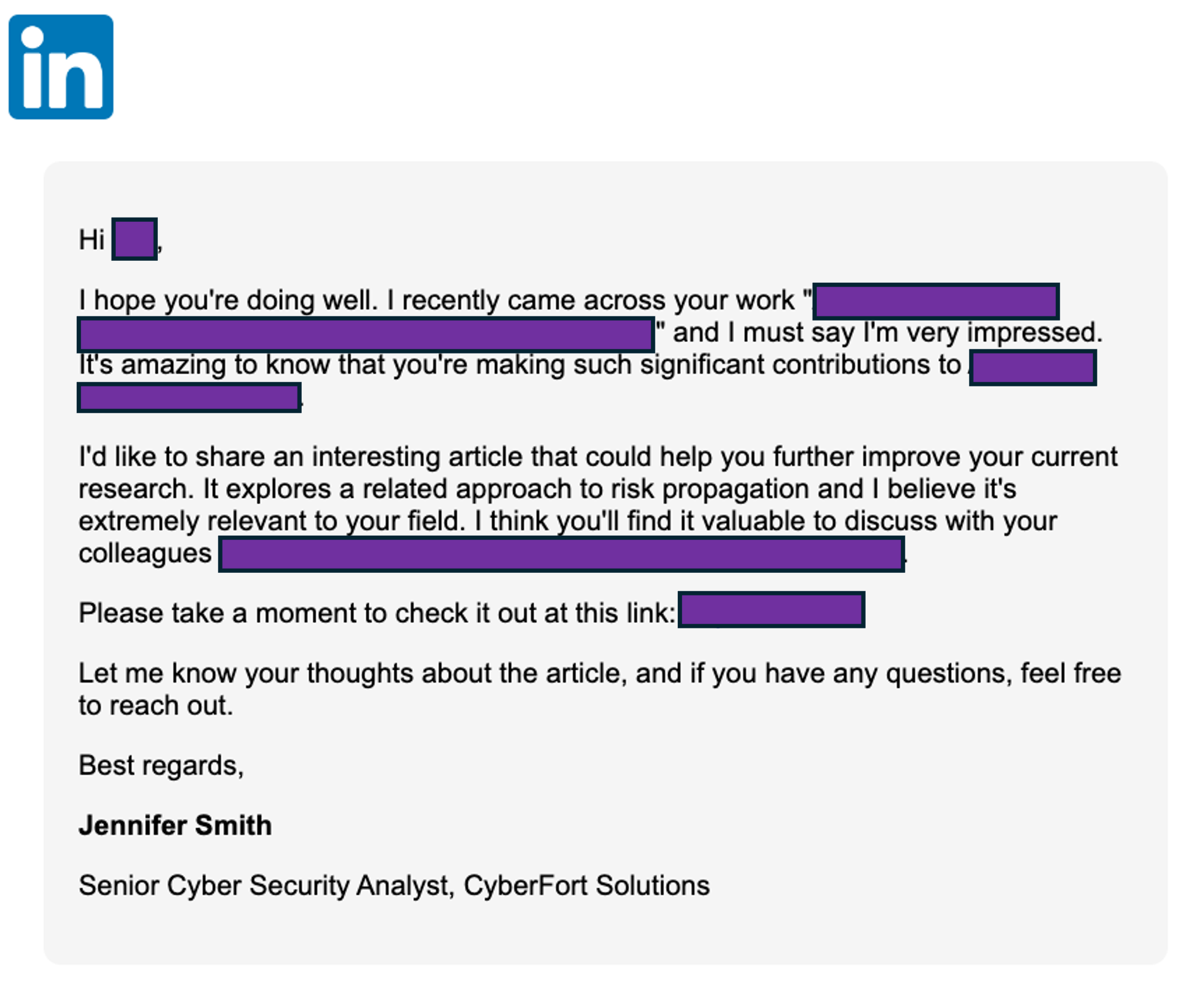}
    \caption{Example of a spear-phishing email message generated by the system (PII blacked out).}
    \label{fig:aspg_example_1}
\end{figure}

To anonymize the dataset for public release, we extracted a subset of the knowledge graph, replaced all PII with fictitious names, and used this anonymized data to generate the 333 spear-phishing messages included in the dataset.
Figure~\ref{fig:aspg_example_1} provides an example of the output produced by the system. 
We anticipate that the prevalence of spear-phishing will surpass commonplace, low-level general phishing campaigns, given the rapid improvement and increased accessibility of open-source LLMs~\cite{hfleaderboard}. 
To the best of our knowledge, there is currently no open-source dataset containing this type of targeted, LLM-generated spear-phishing emails.

\subsubsection{Dataset Specification}
Our dataset consists of 2183 ham (benign) emails and 3317 traditional phishing emails. 
These ham emails were collected from the Enron~\cite{klimt2004enron} and Apache SpamAssassin datasets~\cite{spamassasin}.
The SpamAssassin dataset distinguish between ham and spam emails.
Although spam emails may include phishing content, they are not inherently phishing emails. 
Therefore, we opted to incorporate just the emails classified as hard ham in our benign email collection. 
These hard ham emails predominantly consist of marketing content that bears similarities to phishing emails. 
We assert that evaluating our method using this email category is the most effective means of assessing false positives.
The benign emails collected from the Enron dataset are similar to normal communication emails in a business organization.
The non-spear-phishing emails (i.e., traditional phishing emails) were collected from~\cite{anonymous_2023_8339691}. 
These emails are actual phishing messages dated from 1998 to 2022.
Details on the overall dataset are summarized in Table~\ref{tab:dataset}.
\begin{table}[htbp]
    \centering
    \caption{Dataset description}
    \begin{tabular}{l r}
        \hline
        \textbf{Class} & \textbf{Number of Instances} \\
        \hline
        Phishing~\cite{anonymous_2023_8339691} & 3317 \\
        Enron Ham~\cite{klimt2004enron} & 1781 \\
        Hard Ham~\cite{spamassasin} & 402 \\
        Spear-Phishing & 333 \\
        \hline
    \end{tabular}
    \label{tab:dataset}
\end{table}

\subsection{Feature Space Visualization}
As stated in Section \ref{subsec:concept_drift}, we aim to mitigate concept drift by engineering a feature space $X$ that effectively captures the fundamental characteristics of social engineering attacks. 
To evaluate whether the proposed prompted contextual vectors successfully distill these invariant malicious properties, we employ the t-SNE~\cite{van2008visualizing} dimensionality reduction technique for visualization and analysis of the learned representations.

Dimensionality reduction techniques are commonly used to generate two-dimensional scatter plots of high-dimensional vectors, making it easier to visually distinguish between different target classes in a reduced feature plane. t-SNE (t-Distributed Stochastic Neighbor Embedding) is particularly effective for visualizing high-dimensional data because it preserves the local structure, ensuring that points close together in the high-dimensional space remain close in the lower-dimensional representation.

In a visualization of the optimal feature space $X$, we expect the points representing spear-phishing and traditional phishing emails to be indistinguishable. This would suggest that the vectors are successfully capturing the core malicious intent, irrespective of the stylometric differences. Additionally, we expect these points to be clearly separable from those representing benign samples.

\begin{figure}[t]
    \centering
    \includegraphics[trim={1.8cm 1.1cm 1.8cm 1.1cm},clip,width=\columnwidth]{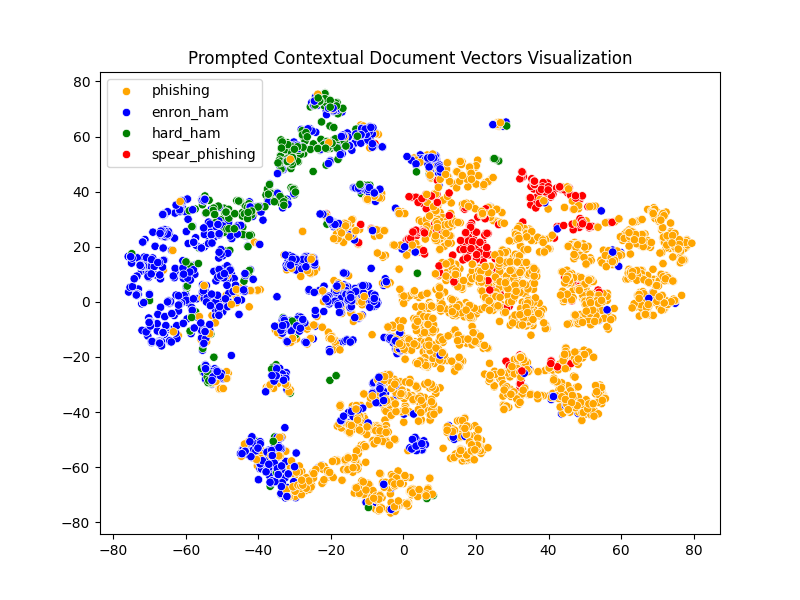}
    \caption{t-SNE visualization - prompted contextual vectors.}
    \label{fig:tsne_gpt}
\end{figure}

Figure~\ref{fig:tsne_gpt} presents the t-SNE visualization of our proposed prompted contextual document vectors. 
In this figure, we can see that our spear-phishing emails (red) are incorporated in the traditional phishing cluster (orange). 
This supports our hypothesis that our vectors manage to capture the core malicious intent in the emails content.

\begin{figure}[t]
    \centering
    \includegraphics[trim={1.8cm 1.1cm 1.8cm 1.1cm},clip,width=\columnwidth]{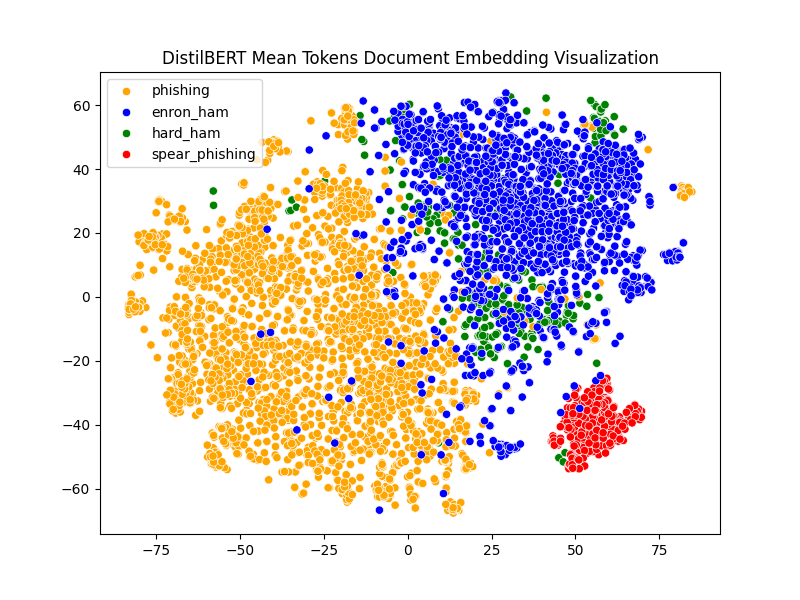}
    \caption{t-SNE visualization - DistilBERT vectors.}
    \label{fig:tsne_distilbert}
\end{figure}

In contrast, in Figure~\ref{fig:tsne_distilbert}, which visualizes the vectors calculated from the mean tokens' embeddings generated from the DistilBERT~\cite{distilbert} transformer model, we can see that the spear-phishing vectors are clustered separately from the traditional phishing samples, with a few 'ham' samples also present in the same cluster. We can assume that spear-phishing emails are grouped together based on stylometric resemblance rather than the contextual malicious intent. This implies that even if a downstream classifier is trained to effectively recognize these emails, it will experience concept drift when encountering new spear-phishing emails produced by a human or a different LLM that generates emails with a different stylometric characteristics.

\begin{table*}[t]
  \centering
  \caption{Spear-phishing detection experiment results}
  \begin{tabular}{lccccc}
    \toprule
    \textbf{Vectors} & \textbf{Recall} & \textbf{Precision} & \textbf{F1} & \textbf{Geometric Mean} & \textbf{False Positive Rate (FPR)} \\
    \midrule
    Prompted Contextual Vectors & 0.96 & 0.87 & \textbf{0.91} & 0.95 & 0.05 \\
    DistilBERT Mean Tokens~\cite{distilbert_mean_tokens} & 0.63 & 0.7 & 0.67 & 0.76 & 0.09 \\
    DistilRoberta~\cite{distilroberta} & 0.13 & 0.65 & 0.21 & 0.35  & 0.02 \\
    MiniLM~\cite{MiniLM} & 0.41 & 0.86 & 0.55 & 0.63 & 0.02 \\
    MPnet~\cite{mpnet} & 0.04 & 0.48 & 0.07 & 0.19 & 0.01 \\
    ada-002~\cite{text_ada} & 0.38 & 0.91 & 0.53 & 0.61 & 0.01 \\
    CountVectorizer~\cite{countvectorizer} & 0.36 & 0.64 & 0.46 & 0.58 & 0.07 \\
    Prompted Contextual Vectors - single question & 0.3 & 0.77 & 0.43 & 0.54 & 0.03 \\
    \bottomrule
  \end{tabular}
  \label{tab:experiment_results}
\end{table*}

\section{Results}
\subsection{Spear-Phishing Detection}\label{subsec:main_experiment}
In this experiment, our objective was to evaluate the effectiveness of the proposed contextual document vectors in detecting LLM-generated spear-phishing emails. 

To formalize the classification process, we considered all of the generated spear-phishing emails (a total of 333) as the test set. 
In the test set, we also include 999 randomly selected benign emails from the Enron and SpamAssassin hard ham collections, proportionally to their distribution in the entire dataset, to enable the measurement of a false positive rate. 

This experimental setup simulates concept drift in the form of covariate shift~\cite{covariate_shift}, where phishing emails evolve from simple, generic scams to sophisticated, AI-generated spear-phishing messages crafted using compromised corporate data.

We anticipate that a weak learner ($k$NN) employing the prompted contextual vectors will identify spear-phishing emails based on their similarity to traditional phishing emails within the feature space. This approach suggests improved alignment with an optimal feature space, $X$, that effectively captures the malicious intent inherent to the phishing class.
We also performed this experiment using document vectors obtained from the top-performing~\cite{sbert} fine-tuned document embedding models available in the Sentence-Transformers library~\cite{sbert}. Specifically, we test the fine-tuned version of the following models: DistilRoberta~\cite{distilroberta}, MiniLM~\cite{MiniLM}, MPnet~\cite{mpnet}. 
These models are trained across multiple domains - question answering, paraphrase mining, and semantic search - producing high-level semantic embeddings. Additionally, we run this experiment using the document embedding model 'text-embedding-ada-002'~\cite{text_ada} available at OpenAI.
We also compared our approach with prompted contextual vectors derived from a single question, where each LLM in the ensemble was tasked with directly classifying the input email.
As an additional baseline, we also compared our method against the term frequency vectorizer from scikit-learn~\cite{countvectorizer}, which has been commonly used in previous phishing detection studies.
Table~\ref{tab:experiment_results} presents the experiment results.

The results show a significant improvement in the recall and F1 score of the kNN classifier when using prompted contextual vectors compared to those obtained from other document embedding models. Given the considerable difference in the recall scores for the vectors produced by our vectorization method and alternative document vectorization methods, we conclude that the prompted contextual vectors demonstrate effectiveness in the task of detecting novel spear-phishing samples.

Interestingly, the findings suggest that vectors generated by more advanced document embedding models are less informative for this particular task than those created by the relatively older DistilBERT mean tokens embedding model. We hypothesize that this result stems from the fact that advanced models are specifically fine-tuned to detect semantic similarity between pairs of sentences or documents~\cite{reimers2019sentencebert}. Although more advanced embedding models typically surpass traditional embedding models in numerous NLP tasks~\cite{sbert}, their fine-tuning method can become a disadvantage in adversarial problem domains like spear-phishing detection. In such cases, the samples we aim to identify are actively attempting to preserve semantic similarity to benign samples, making detection more challenging.

It is important to note that we chose the $k$NN classifier, often regarded as a weak learner, to thoroughly assess the quality of prompted contextual vectors against other document vectorization methods. In practical applications, opting for a stronger model would be advantageous. Such a model could benefit significantly from hyperparameter tuning and decision threshold optimization to achieve optimal results that align closely with user requirements.
To demonstrate this, we also trained a CatBoost classifier in the same experimental setting, achieving better results of 0.99 F1 score and 0.01 FPR. The same results were achieved by training an XGBoost classifier.

\subsection{General Phishing Detection}
To validate our detection method's ability to detect general phishing emails and evaluate its overall robustness, we conducted a five-fold cross-validation classification experiment as follows.
Initially, we created a distinct holdout dataset that included just the 333 generated spear-phishing emails, ensuring that this dataset remains part of the test set throughout all folds.
Then, we conducted a standard stratified five-fold cross-validation experiment, using different email types (Enron ham, SpamAssasin hard ham, and phishing) as labels.
However, the actual classification focused on distinguishing between benign and phishing labels for each email.

Essentially, the training and test splits included all types of emails in a stratified manner, except for spear-phishing emails, which are present exclusively in the test splits.
We also performed this experiment using the second-best-performing embedding model from the previous experiment - the DistilBERT mean tokens embedding model.
\begin{table*}[t]
  \centering
  \caption{General phishing detection: mean performance metrics across folds}
  \begin{tabular}{lccccc}
    \toprule
    \textbf{Vectors} & \textbf{Mean Recall} & \textbf{Mean Precision} & \textbf{Mean F1 Score} & \textbf{Mean G-Mean} & \textbf{Mean FPR} \\
    \midrule
        Prompted Contextual Vectors & 0.95 & 0.99 & \textbf{0.97} & 0.96 & 0.03 \\
        DistilBERT Mean Tokens & 0.86 & 0.97 & 0.92 & 0.9 & 0.05 \\
    \bottomrule
  \end{tabular}
  \label{tab:exp3_metrics}
\end{table*}
Table~\ref{tab:exp3_metrics} presents the mean values of the metrics observed across all folds. 
It is noteworthy that all metrics exhibit improvement compared to the previous experiment. Additionally, we observed a narrower performance gap between the two vectorization methods compared to the previous experiment.
This outcome is logical, since the previous experiment tackled a more challenging task, focusing on the detection of spear-phishing emails exclusively based on regular phishing and benign samples.
In contrast, the current experiment includes regular phishing samples in the test sets of all folds.

\subsection{General Social Engineering Detection}
In order to evaluate our approach ability to generalize to other social engineering attacks, we examine its performance on a dataset~\cite{timko2024smishing} comprised of smishing~\cite{ibm_smishing} messages. 
From this dataset, we selected 182 smishing messages containing links that were confirmed to be malicious by VirusTotal (with a confidence score greater than 5).
Additionally, we collected 581 benign SMS messages from the UCI SMS corpus~\cite{sms_dataset_paper}.
The classification experiment was conducted using a training dataset that exclusively contained email phishing, spear phishing, and ham (non-phishing) messages, while the test set consisted solely of the selected SMS messages. 
SMS messages differ significantly from emails in length and writing style, introducing a semantic shift that a general social engineering detection method must navigate.
In this experiment we compare the prompted contextual vectors against vectors generated by the DistilBERT mean tokens embedding model.
Table~\ref{tab:smishing_results} displays the classification results obtained using both vectorization methods. 
The experimental results demonstrate that when using the same $k$NN classifier, prompted contextual vectors significantly outperform DistilBERT mean token embeddings in handling semantic drift of social engineering messages across different communication mediums. This performance gap provides empirical support for our hypothesis that prompted contextual vectors achieve better alignment with the theorized optimal feature space $X$, successfully capturing underlying malicious intent rather than basic semantic features.
\begin{table*}[t]
\centering
\caption{Social engineering generalization experiment result - smishing dataset}
\begin{tabular}{lccccc}
\hline
\textbf{Vectors} & \textbf{Recall} & \textbf{Precision} & \textbf{F1} & \textbf{Geometric Mean} & \textbf{FPR} \\
\hline
Prompted Contextual Vectors & 0.93 & 0.88 &\textbf{0.9} & 0.94 & 0.04  \\
DistilBERT Mean Tokens & 0.49 & 0.34 & 0.4 & 0.59 & 0.3 \\
\hline
\end{tabular}

\label{tab:smishing_results}
\end{table*}

\subsection{Small Language Models Ensemble}
In this experiment, we maintained the same settings as described in section \ref{subsec:main_experiment}, but replaced the original LLM ensemble with smaller, open-source models. Specifically, we used Llama 3.1 8B~\cite{llama31_8b}, Llama 3 8B~\cite{llama3_7b}, Phi 3 Medium~\cite{phi3medium}, and Mistral Nemo~\cite{mistral_nemo}. This configuration yielded a recall score of 0.92, precision of 0.78, an F1 score of 0.85, G-Mean score of 0.93, and FPR of 0.08, which were inferior to the original ensemble’s performance when paired with $k$NN as the downstream classifier.

However, when we repeated the experiment using an Extra Trees classifier with an optimized decision threshold, it resulted with a recall of 0.89, precision of 0.9, an F1 score of 0.89, and a G-Mean score of 0.93, while maintaining a false positive rate of 0.03.
This suggests that while smaller LLMs produce less informative prompted contextual vectors, they can still be effectively leveraged with a more powerful downstream classifier.

\subsection{LLM Ensemble Ablation Study}
To assess the contribution of each LLM in the ensemble, we conducted an ablation study by running the main experiment with every possible combination of LLMs. Table~\ref{tab:ablation_results} presents the results of this study.
The findings indicate that Gemini Pro makes the most significant contribution to the ensemble. Notably, even when generating prompted contextual vectors using Gemini exclusively, it achieves an impressive F1 score of 82\%.
The best-performing combination is Gemini Pro and GPT-4, achieving similar results to the main experiment using the entire ensemble (as outlined in the main experiment).

This ablation study demonstrates the value of utilizing an ensemble of LLMs in our approach. While individual models show varying strengths, their combinations consistently outperform single-model configurations.
It is important to note that the main experiment is performed using the kNN classifier to demonstrate the effectiveness of the prompted contextual vectors. Using a stronger classifier in the detection pipeline could result in better performance for different configurations, as the classifier can learn the contribution of every LLM in the ensemble and better optimize their weights in the final classification.

\begin{table*}[t]
  \centering
  \caption{LLM ensemble ablation study results}
  \begin{tabular}{lccccc}
    \toprule
    \textbf{LLMs Ensemble} & \textbf{Recall} & \textbf{Precision} & \textbf{F1} & \textbf{Geometric Mean} & \textbf{False Positive Rate (FPR)} \\
    \midrule
    GPT 4, Gemini Pro & 0.98 & 0.85 & 0.91 & 0.96 & 0.06 \\
    Gemini Pro, GPT 3.5 & 0.90 & 0.83 & 0.86 & 0.92 & 0.06 \\
    Gemini Pro & 0.94 & 0.73 & 0.82 & 0.91 & 0.12 \\
    GPT 4 & 0.95 & 0.66 & 0.78 & 0.89 & 0.16 \\
    GPT 4, GPT 3.5 & 0.92 & 0.68 & 0.78 & 0.89 & 0.14 \\
    GPT 3.5 & 0.75 & 0.60 & 0.66 & 0.79 & 0.17 \\
    \bottomrule
  \end{tabular}
  \label{tab:ablation_results}
\end{table*}

\subsection{Questions Ablation Study}
The effectiveness of prompted contextual vectors inherently depends on the quality of the questions used to generate them. For this study, we manually crafted questions specifically tailored to spear-phishing detection, as detailed in section \ref{subsec:questions}. Given the manual nature of this process, without systematic optimization, it is reasonable to assume that our question set may not be optimal.
To quantify the contribution of each individual question to the classification performance, we conducted a systematic ablation study. Following the experimental setup described in \ref{subsec:main_experiment}, we performed seven iterations of the experiment, each time removing one question from the set. For each iteration, we measured the F1 loss—the difference between the original experiment's F1 score and the F1 score obtained with the reduced question set. A positive F1 loss indicates the removed question's contribution to the overall classification performance, with larger values suggesting greater importance of that particular question.

\begin{figure}
    \centering
    \includegraphics[clip,width=\columnwidth]{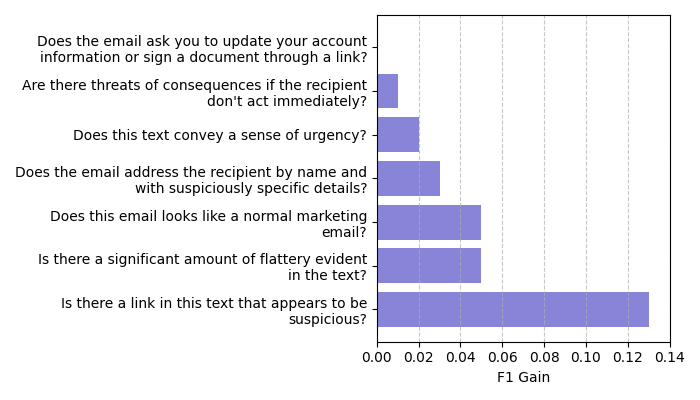}
    \caption{Questions ablation study results.}
    \label{fig:questions_ablation_study}
\end{figure}

The results of this ablation study are presented in Figure \ref{fig:questions_ablation_study}. The analysis reveals that the question probing for suspicious links provides the strongest signal, with an F1 loss of 0.13. The remaining questions also demonstrate meaningful contributions to the classification performance, with F1 losses ranging from 0.01 to 0.05. Notably, only one question showed no impact on performance: "\textit{Does the email ask the recipient to update an account information or sign a document through a link?}" This finding suggests this particular question could be removed from the question set without degrading classification accuracy.

\subsection{Error Analysis and Explainability}
Due to the inherent interpretability of the prompted contextual vectors, a comprehensive error analysis can be performed to enhance detection pipelines utilizing these vectors. 
In this section, we describe the analysis we performed in the context of the main experiment described in section \ref{subsec:main_experiment}. 
For this analysis, we selected a randomly misclassified spear-phishing sample (erroneously classified as a benign sample) and a random misclassified benign sample. 
To perform the error analysis, we calculated the standard deviation of every answer to every question across the LLM ensemble, aiming to identify disagreement between the models.

\begin{figure}[h]
    \centering
    \includegraphics[width=0.99\linewidth]{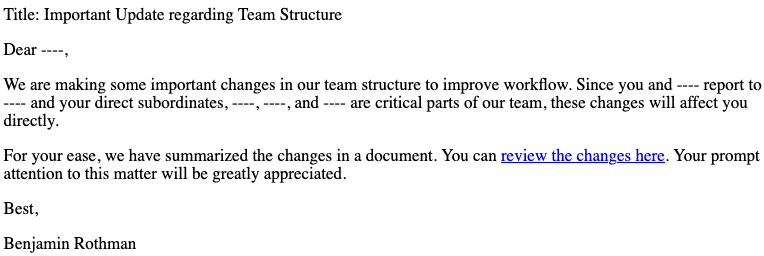}
    \caption{Misclassified spear-phishing sample (PII removed).}
    \label{fig:misclassified_spear_phishing}
\end{figure}

Figure \ref{fig:misclassified_spear_phishing} presents a misclassified spear-phishing sample. 
This sample will be hard to detect, as it appears to be a legitimate business conversation without any obvious presence of a commonly used persuasion technique. 
When examining the prompted contextual vector of this sample, we noticed high disagreement between the LLMs for the question: \textit{Is there a link in this email that appears to be suspicious?}. 
The link embedded in the text is a short URL which represents the typical output of an URL shortening service. These services are frequently used by attackers to disguise a malicious link. 
Upon further examination of the prompted contextual vector of this sample, we found that the output of every LLM regarding this question is as follows: GPT-3.5: 0, GPT-4: 0.5, Gemini Pro: 0.8. From these outputs we can infer that GPT-3.5 does not deem such links as suspicious, GPT-4 expresses uncertainty, and Gemini Pro leans toward being suspicious of these types of links. 

\begin{figure}[h]
    \centering
    \includegraphics[width=0.99\linewidth]{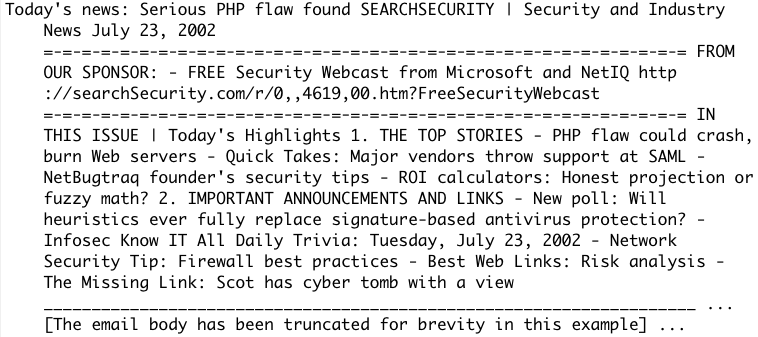}
    \caption{Misclassified hard ham sample.}
    \label{fig:misclassified_benign}
\end{figure}

Figure~\ref{fig:misclassified_benign} presents a misclassified benign sample that belongs to the hard ham category in the SpamAssasin~\cite{spamassasin} dataset. In our examination of this sample, we observed high disagreement among the LLMs for the following questions:
\begin{itemize}
    \item \textit{Does this email convey a sense of urgency?}
    \item \textit{Is there a link in this email that appears to be suspicious?}
    \item \textit{Does this email look like a marketing email?}
\end{itemize}

\noindent Interestingly, this sample appears suspicious despite being labeled as a benign marketing email. 
Both GPT-3.5 and Gemini Pro assigned a high value to the question inquiring if the email conveys a sense of urgency. 
This is attributed to the email's content, which discusses a PHP security flaw that could potentially "crash and burn" web servers. 
Among the ensemble, Gemini Pro was the only LLM that assigned high values to the questions asking whether the email resembles a marketing email and if it contains any suspicious links.

Various approaches can be used to avoid such misclassifications:

\begin{enumerate}
    \item Impute the output probabilities of models that underperform for a specific question.
    \item Trace back the reasoning steps of every underperforming model, identify pitfalls, and refine the existing question or formulate a new question to compensate for these reasoning pitfalls.
\end{enumerate}

In summary, the inherent explainability  of the prompted contextual vectors allows for a thorough analysis of errors and effective mitigation. This capability is significantly limited in previously proposed feature vectors and vectors produced by other document vectorization methods.

\subsection{Comparison with Established Detection Methods}
As mentioned in the literature review outlined in section~\ref{subsec:phishing_related}, the most recent and comprehensive phishing detection study we surveyed was the work of Bountakas et al.~\cite{bountakas2023helphed}. Unfortunately, the authors did not publish their method's code, making it difficult to replicate all their manually engineered features. Moreover, their method relies on data beyond the email content's text (e.g., header, attachments), in contrast to our method, which relies solely on the email's text.
However, when the authors compared their method with other state-of-the-art approaches, they found that the work of Gualberto et al.~\cite{gualberto2020answer}, which relies only on the email text, reported slightly superior results with a perfect F1 score.

Gualberto et al. also did not publish their code, so we made our best effort to replicate their proposed method and evaluate it in the main experiment outlined in section~\ref{subsec:main_experiment}. We implemented the text preprocessing function according to their specifications using the NLTK~\cite{bird2009natural} Python library, and the feature dimensionality reduction, feature selection, and classification pipelines using scikit-learn~\cite{pedregosa2011scikit}.

\begin{figure}
    \centering
    \includegraphics[width=1\linewidth]{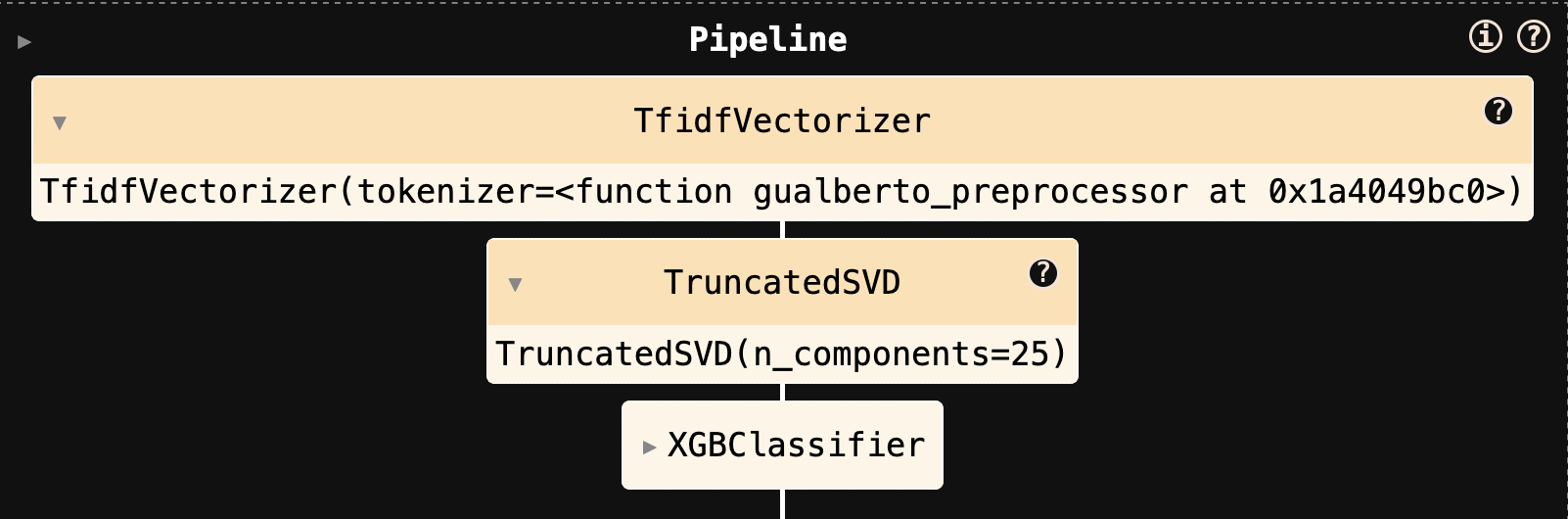}
    \caption{Implemented pipeline of the first method suggested by Gualberto et al.}
    \label{gualberto1}
\end{figure}

\begin{figure}
    \centering
    \includegraphics[width=1\linewidth]{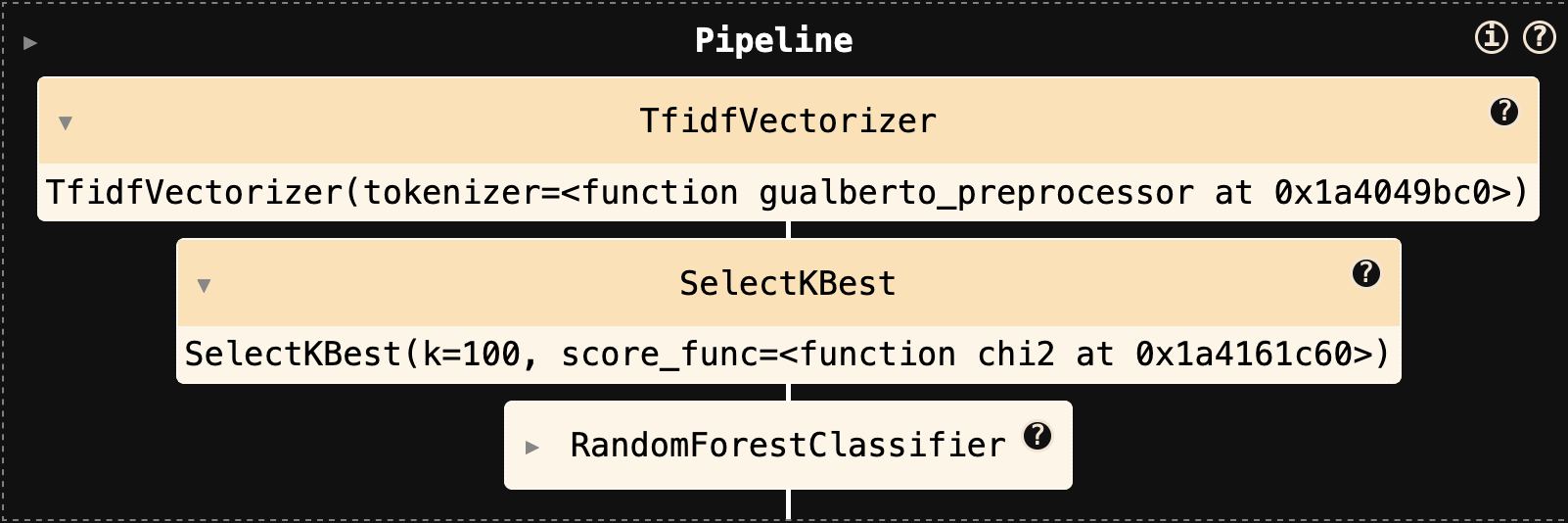}
    \caption{Implemented pipeline of the second method suggested by Gualberto et al.}
    \label{gualberto2}
\end{figure}

Gualberto et al. proposed two distinct detection methods, both employing a TF-IDF representation of the email’s textual content with stop words and punctuation removed and applying lemmatization. The first method conducts feature dimensionality reduction using LSA, resulting in 25 features fed into an XGBoost classifier. The second method employs Chi-square feature selection, retaining 100 features for a Random Forest classifier. Figures~\ref{gualberto1} and ~\ref{gualberto2} outline the implemented pipelines of the first and second methods, respectively.

Table~\ref{tab:sota_compare} presents the performance of both methods in the primary experiment. Notably, neither method's performance is comparable to our suggested approach, even though our approach employs a simple kNN classifier without any preprocessing steps. This observation underscores the effectiveness of using prompted contextual vectors, which capture the malicious intent within the text rather than relying on the frequency of specific words.

\begin{table*}[t]
  \centering
  \caption{Performance of the methods proposed by Gualberto et al. in the main experiment}
  \begin{tabular}{lccccc}
    \toprule
    \textbf{Detection Method} & \textbf{Recall} & \textbf{Precision} & \textbf{F1} & \textbf{Geometric Mean} & \textbf{False Positive Rate (FPR)} \\
    \midrule
    Gualberto et al. - Method 1 & 0.23 & 0.92 & 0.37 & 0.48 & 0.01 \\
    Gualberto et al. - Method 2 & 0.5 & 0.92 & 0.65 & 0.7 & 0.02 \\
    \bottomrule
  \end{tabular}
  \label{tab:sota_compare}
\end{table*}

\section{Discussion}

This paper introduces a novel document vectorization method, which utilizes the reasoning capabilities of an ensemble of LLMs to vectorize a document. We employ this vectorization method for the task of detecting targeted, personalized, LLM-generated spear-phishing emails. Previous studies that used ML for spear-phishing detection have mainly taken the attribution approach, aiming to classify a new email to a known author or spear-phishing campaign in order to perform detection. However, this approach is not effective against unknown spear-phishing campaigns. Moreover, none of the prior studies evaluated their approach against LLM-generated spear-phishing emails.

The recent surge in AI advancements has enabled the creation of increasingly sophisticated social engineering attacks~\cite{zvelo_llm_recon}, making it likely that such attacks will become more prevalent in the near future. Our literature review reveals that recent phishing detection studies and proposed methods have predominantly been trained and evaluated on traditional, outdated phishing messages. This suggests that many current detection solutions may be ill-equipped to identify AI-generated phishing and spear-phishing attacks, as they often rely on basic features like word counts and specific keywords~\cite{gutierrez2018learning,salloum2021phishing}.

In the era of LLMs, phishing detectors should be robust to semantic shifts, since different prompting techniques and LLM variations can produce phishing attacks that deviate significantly from the stylometric features present in the training data. To address this, a modern detector could benefit from leveraging an optimal-aligned feature space that captures malicious intent rather than stylometry. This would increase resilience against the rapid dataset shifts brought about by the fast-paced development and deployment of malicious AI tools~\cite{fbi_ai_phishing_warning}.

Our evaluation begins with a comparative visualization of two feature spaces: the prompted contextual vectors and the embeddings generated by the DistilBERT model. The visualization reveals that the prompted contextual vectors' feature space aligns more closely with the theorized optimal feature space $X$, as evidenced by the indistinguishable distribution of traditional phishing and AI-generated spear-phishing samples. Conversely, in the DistilBERT feature space visualization, these samples form distinct clusters, suggesting that models trained on either category would likely suffer from concept drift when attempting to classify the other.

To validate this observation, we conducted classification experiments specifically designed to test robustness against concept drift. We constructed our experimental setup with a clear division: the test set exclusively contained spear-phishing messages, while the training set comprised traditional phishing and benign samples. Notably, a simple $k$NN classifier, leveraging the prompted contextual vectors as input, demonstrated robust performance despite this intentional covariate shift, with an F1 score of 0.91. This is particularly significant as $k$NN operates by directly comparing input vectors to its reference samples, rather than learning complex decision boundaries or feature transformations like other ML models. The strong performance suggests that the prompted contextual vectors effectively capture the fundamental characteristics of malicious intent across different phishing variants. 
When we tested alternative vectorization methods as input for the $k$NN classifier, none were able to provide effective classification results under our experimental settings, further highlighting the unique capability of prompted contextual vectors in this domain.

To further assess the alignment between prompted contextual vectors and the optimal feature space $X$, we designed an experiment with an extreme semantic shift: training on email samples (both malicious and benign) and testing on SMS messages (smishing and benign). While both phishing and smishing share the fundamental goal of enticing targets to interact with malicious links or attachments, they exhibit distinct linguistic characteristics. SMS messages are typically shorter and employ a more informal tone compared to emails. Despite these substantial stylistic differences, the $k$NN classifier using prompted contextual vectors achieved an F1 score of 0.9 on the SMS test set. In contrast, the DistilBERT embeddings, which performed best among our baselines in previous experiments, only achieved an F1 score of 0.4. These results strongly support our hypothesis that prompted contextual vectors capture the underlying malicious intent rather than surface-level linguistic features.

While this study focused on spear-phishing detection, the application of prompted contextual vectors extends beyond this specific domain. The method's core strength lies in its use of human-crafted questions that prompt LLMs to reason about contextual aspects of documents. This approach is particularly promising for adversarial problem domains where traditional feature extraction methods may fall short, such as fraud detection and LLM input moderation. In these domains, like in spear-phishing detection, the ability to capture underlying intent rather than surface-level features is essential for robust classification.

Research in spear-phishing detection remains limited, despite its importance. The emergence of LLMs has fundamentally transformed the threat landscape, creating a need for suitable detection methods. To accelerate research in this domain, we are making two key contributions available to the research community: (1) our dataset of LLM-generated spear-phishing emails, created using our collaborators' proprietary system, and (2) our complete code repository, including the implementation of prompted contextual vectors and all prompts used in our experiments.

\subsection{Privacy}
A real-world phishing detection system that incorporates prompted contextual vectors as part of its input should avoid using public LLM APIs due to significant privacy and security concerns. Transmitting potentially sensitive email content or user data to external servers could compromise confidentiality and expose organizations to data breaches. Moreover, there's a risk that emails processed through public APIs might be used as training data for future iterations of the LLM, potentially exposing confidential information to a wider audience.

We evaluated our detection method utilizing OpenAI models served from a private, secure Azure OpenAI instance. All prompts sent and completions received from LLMs in an Azure OpenAI instance are not available to OpenAI and are not used to improve any Microsoft or third-party products or services\footnote{https://learn.microsoft.com/en-us/legal/cognitive-services/openai/data-privacy}. To access Gemini Pro, we did use a public API. However, a secured private API is available via Google Cloud\footnote{https://cloud.google.com/gemini/docs/discover/data-governance}. Furthermore, accessing OpenAI models securely is also possible by using ChatGPT Enterprise\footnote{https://openai.com/chatgpt/enterprise/}.

Our proposed vectorization method is not limited to closed-source models. It is also possible to use it by leveraging locally-hosted open-source models. Open-source models are becoming increasingly capable and are often a popular choice for powering privacy-sensitive applications, offering greater control over data handling and customization options to meet specific security requirements.

\subsection{Limitations}
Our proposed vectorization method has a few limitations worth mentioning:
First, unlike other document vectorization methods, our method contains several configurable elements. 
The most prominent of these elements are the questions posed to the LLMs. 
The quality of the vectors generated using this method hinges on the quality of the devised questions. 
In our experiments, we manually crafted these questions based on a literature review of prevalent persuasion tactics used in social engineering.
This process can be quite time-consuming. 
Furthermore, the selection of LLMs to incorporate into the ensemble also have an impact on downstream performance. 
However, building an ML pipeline often involves manual trial-and-error experimentation, whether it be feature engineering and selection, model selection, hyperparameter tuning, or other aspects.

\subsection{Future Work}
While prompted contextual vectors demonstrate promising results, their current implementation relies on manually crafted questions, presenting an opportunity for improvement. Although our question set proved effective for spear-phishing detection, we believe there is potential to better align these questions with an optimal feature space for social engineering detection. Our primary direction for future research is the development of a domain-agnostic methodology for automatic question generation. Such a framework would not only reduce the manual effort required but could also discover more effective questions than those devised through human intuition. This advancement would enhance the applicability of prompted contextual vectors across diverse domains, making the method more robust and easier to adapt to new classification tasks.

\bibliographystyle{IEEEtran}  
\bibliography{ref}

\appendices

\section{\label{sec:availability}Data Availability}
The implementation code for the vectorization method, along with the spear-phishing emails generated and the prompts utilized, can be accessed in the provided \href{https://github.com/nahmiasd/Prompted-Contextual-Vectors-for-Spear-Phishing-Detection}{repository}. 
The 'data' folder contains the collection of spear-phishing emails generated.
\section{Automatic Spear Phishing Generator - Overview}

\section{Comparison With Established Methods - Implementation Details}

\end{document}